\documentclass[11pt,a4paper]{article}
\usepackage[table]{xcolor}

\usepackage[hyperref]{acl2021}
\usepackage{times}
\usepackage{latexsym}

\usepackage{booktabs}
\usepackage{multirow}
\usepackage{makecell}
\usepackage{threeparttable}

\usepackage{graphicx}
\usepackage{amsmath}

\usepackage{microtype}

\usepackage[inline]{enumitem}

\aclfinalcopy 

\title{Grounding and Distinguishing Conceptual Vocabulary Through Similarity Learning in Embodied Simulations}

\author{Sadaf Ghaffari \and Nikhil Krishnaswamy \\
  Situated Grounding and Natural Language (SIGNAL) Lab \\
  Department of Computer Science, Colorado State University \\
  Fort Collins, CO, USA \\
  \texttt{\{sadafgh,nkrishna\}@colostate.edu}}

\date{}

\begin{document}
\maketitle
\begin{abstract}
We present a novel method for using agent experiences gathered through an embodied simulation to ground contextualized word vectors to object representations. We use similarity learning to make comparisons between different object types based on their properties when interacted with, and to extract common features pertaining to the objects' behavior.  We then use an affine transformation to calculate a projection matrix that transforms contextualized word vectors from different transformer-based language models into this learned space, and evaluate whether new test instances of transformed token vectors identify the correct concept in the object embedding space. Our results expose properties of the embedding spaces of four different transformer models and show that grounding object token vectors is usually more helpful to grounding verb and attribute token vectors than the reverse, which reflects earlier conclusions in the analogical reasoning and psycholinguistic literature.
\end{abstract}

\vspace*{-2mm}
\section{Introduction}
\label{sec:intro}
\vspace*{-2mm}

A common critique of modern large language models (LLMs) is that they lack {\it understanding} in the sense of being able to link an utterance to a specific communicative intent~\cite{bender-koller-2020-climbing}. This shortcoming is often characterized as being due to a lack of ability to {\it ground} or link lexical items to real-world entities such as classes of objects, or associated properties or actions. For instance, a modern generative LLM like ChatGPT\footnote{\url{https://chat.openai.com}} may be able to generate coherent text describing an object (e.g., ``a \textit{coconut} has a hard, often hairy outer shell''), without any inherent underlying conceptualization of what the item actually {\it is}. 

Crucially, these underlying conceptualizations necessarily invoke other modalities.  Existing approaches to grounding in NLP typically treat the problem as one of making the correct kind of link between text and another modality, usually images~\cite{socher-etal-2014-grounded,yatskar2016situation,zhu-etal-2020-multimodal,zhu-etal-2021-multimodal}.  However, still images do not capture the wealth of information humans receive when interacting with objects or experiencing events, and video data requires orders of magnitude more data and computational power to effectively process.  Additionally, humans do not use vision alone as their only non-linguistic modality.

As humans develop object concept representations and map them to associated nouns, they are also learning to individuate objects from the perceptual flow not just based on visual features but also based on experience that includes interacting with them in real time~\cite{spelke1985perception,spelke1989object,spelke1990principles,baillargeon1987object}.  \citet{gentner2006verbs} argues that \citet{talmy1975semantics}'s findings on variability in verbal semantics helped to explain why nouns are typically learned before words for verbs or other properties.  Concrete nouns are more easily ``groundable'' not just because of their visual manifestations but also because of their physical presences that leave traces in the world, and these physical properties provide a scaffold on which to build representations of related concepts that are supervenient upon understanding of objects.

\begin{figure*}
    \centering
    \includegraphics[width=\textwidth]{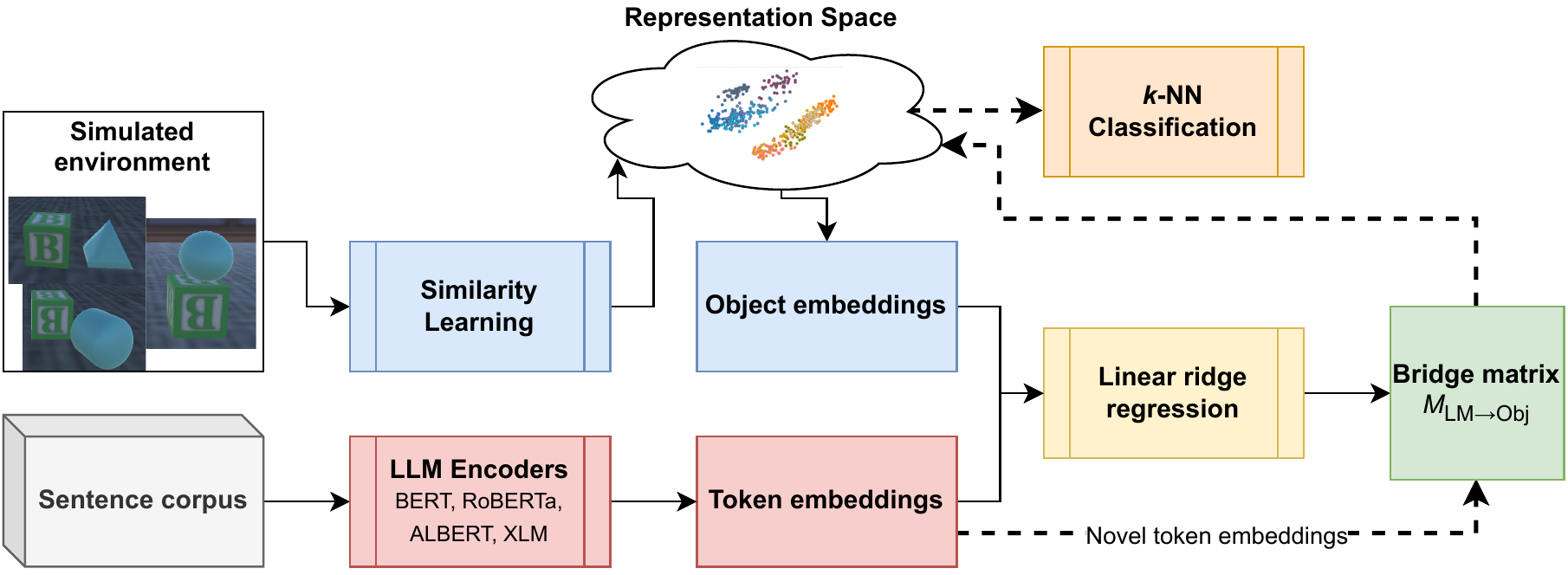}
    \caption{Overview of grounding architecture. In this figure, $M$ denotes the computed affine transformation matrix between language model (LM) and object classifier (Obj) space.  Similarity learning in this figure is performed only over a subset of the available classes (see Sec.~\ref{ssec:similarity}). The solid lines depict the flow of information used to ``train'' or compute the affine transformation ``bridge'' matrix, and the dashed lines depict the flow of information of novel ``test'' samples, including transformation by the precomputed bridge matrix.}
    \label{fig:overview}
\end{figure*}

In this paper we take an {\it embodied simulation} approach to grounding, using a virtual environment to create experiences for an agent interacting with objects. We show that similarity learning over data gathered during the agent's experience in the virtual world can not only make comparisons between objects, but also appears to learn information pertaining to more abstract properties of the objects.  Fig.~\ref{fig:overview} shows a schematic view of our overall approach. We map token vectors from different transformer-based LLMs into the resulting representation space, and show that with just a few samples, grounding noun representations alone is helpful for subsequent grounding of verbal tokens, abstract properties, and attributive terms, but that grounding verbal or attributive token representations is less helpful for subsequent grounding of object concepts.







\vspace*{-2mm}
\section{Related Work}
\label{sec:related}
\vspace*{-2mm}

Multiple works in cognitive science have identified contrastive mechanisms, and the ability to analogize by applying previous experiences to novel scenarios, as a cornerstone of problem-solving~\cite{gentner1983structure, forbus1995mac, mclure2010learning, hofstadter2013surfaces, smith2014role, lovett2017modeling}.

In visual analogy, \citet{hill2018learning} created analogies by contrasting relational structure. For solving Raven’s Progressive Matrices (RPMs)~\cite{james1936raven}, \citet{malkinski2022multi} applied a generalization of the Noise Contrastive Estimation (NCE) algorithm~\cite{gutmann2010noise}. 
\citet{wu2018unsupervised} performed feature learning using visual similarity via unsupervised learning at the instance-level with NCE. \citet{oh2016deep} used deep feature embedding based on lifted structure loss, and evaluated their method via clustering and retrieval tasks on images from unseen classes. \citet{bell2015learning} trained a Siamese CNN with contrastive loss~\cite{hadsell2006dimensionality} to learn an embedding space of interior design images and applied the embeddings to image search applications like finding visually similar products across categories.


Since evaluating AI agents in physical environment can be expensive, many works have used both embodied and non-embodied simulations to explore language learning. \citet{hermann2017grounded} combined reinforcement and unsupervised learning to teach agents to correlate linguistic symbols with physical percepts and action sequences. However, this still-computationally-expensive method required millions of training episodes. The SNARE benchmark~\cite{thomason2022language} was evaluated on grounding to objects but not in context or under interaction. \citet{tucker2021emergent} demonstrated an emergent clustering of semantic tokens from a (non-embodied) continuous representation space and \citet{tucker2022towards} extended that method with an application of an information bottleneck. Our work integrates concepts from the above areas: embodied simulation environments, language grounding using both situated and linguistic context, and emergent semantic categorization.

\citet{merullo2022pretraining} examined 2D and 3D visual and interactive data for learning object affordances and found that 3D and interactive data performed better. \citet{ebert2022trajectories} extracted verbal semantics from object trajectories in 3D space, but focused only on verbs whereas we examine nouns, verbs, and attributes. We show that objects and properties can also be encoded by object trajectories or behavior in 3D space, using a stacking task that exposes richer correlations between object properties and behaviors.
Like us, \citet{patel2022mapping} investigated word grounding but they evaluated on within-domain concepts (e.g., learning ``left'' to help ground ``right'' where we investigate how, say, learning ``sphere'' can help ground ``round'') in a grid world (our world representation is continuous), and where their transformation passed input through a whole LLM, our transformation is a simple affine map between embedding spaces. \citet{lazaridou-etal-2015-combining} mapped vision embeddings to language via ridge regression, but their Multimodal Skip-Gram used static word vectors, not contextualized vectors from transformers.


\citet{pezzelle-etal-2021-word} evaluated the representations of transformer models and found that multimodal representations better align with human judgments in the domain of concrete nouns, but not abstract terms. Our work arrives at a related conclusion using cross-model transfer.


\vspace*{-2mm}
\section{Methodology}
\label{sec:method}
\vspace*{-2mm}

Our methodology comprises two primary components: similarity learning to create a representation space of objects by making comparisons between geometric properties, and linear projection to ground language representations to this space.

\vspace*{-2mm}
\subsection{Data}
\label{ssec:data}
\vspace*{-1mm}

We use the dataset from~\citet{ghaffari2022detecting}, in which an agent in a simulated environment built on the VoxWorld platform~\cite{krishnaswamy-etal-2022-voxworld}, stacks 9 different types of {\it theme} objects\footnote{{\it cube}, {\it sphere}, {\it cylinder}, {\it capsule}, {\it small cube}, {\it egg}, {\it rectangular prism}, {\it pyramid}, and {\it cone}.} on top of a cube.  Each object's behavior when stacked is different, based on its geometric structure and therefore {\it affordances}~\cite{gibson1977theory}.  For instance, a cube, if placed correctly on another cube, will remain stacked, while a sphere placed in the same position will roll off and keep moving. An egg will likely do the same, but the direction of motion may be subtly different based on the symmetry of the object. The dataset contains 10,000 total samples, each with 43 numerical values describing the behavior of the objects in the course of this stacking task: theme object type; object orientation before the agent acts upon it; numerical action describing the placement of the theme relative to the destination object; resulting spatial relations between the two objects; object orientation after the action; and position of the theme relative to the destination object before action, immediately after action, and after the world physics are applied to the scene. See~\citet{ghaffari2022detecting} for further details. This information about object behaviors and trajectories in space, unlike still images, {\it situates} the objects in an embodied environment and encodes richer information than visuals alone do. The dataset does also contain images but these are not used here.

Two of the object types in the data, {\it cylinder} and {\it cone}, have both flat sides and round edges, and as this distinction strongly affects the behavior of these objects when stacked (i.e., given proper placement, a cylinder or cone will stack on top of a cube but only if also placed in the correct orientation), the dataset preserves these distinctions nicely, and we split the cone and cylinder samples into ``flat-side-down'' and ``round-edge-down'' for similarity learning of properties (Sec.~\ref{ssec:similarity}).

\vspace*{-2mm}
\subsection{Similarity Learning of Object Properties} 
\label{ssec:similarity}
\vspace*{-1mm}

Since comparing pairs of examples plays a role in analogy-making, we apply deep pair-based learning to compare structural object properties. The main goal in deep pair-based learning techniques is to learn an embedding space where embeddings of similar samples are closer together and dissimilar samples are pushed apart, after the projection of input space to the embedding metric space. In our case, the trained model should be able to infer contrasts and comparisons between different  structural properties of objects (in this case {\it flatness} and {\it roundness}), and apply it to novel objects based on commonalities in behavior and relational structure. 

In training, we consider only samples of {\it cube}, {\it rectangular prism}, {\it pyramid}, and {\it small cube} that stacked successfully, and samples of {\it capsule}, {\it sphere}, and {\it egg} that did not.  For testing data, we take a test split of the same object classes, and also samples of {\it cone} and {\it cylinder}.  These samples behave differently according to, among other things, their orientation when placed. We split cone and cylinder instances into ``flat-side-down'' (stacked successfully) and ``round-edge-down'' (did not stack successfully). Therefore we train on 7 classes and evaluate on 11 classes, including 4 never seen in training.

To train, we take 500 samples of each training class, zero-center the data and make it unit variance.
Our model architecture consists of 4 1D convolutional layers (32, 32, 64, and 64 units, respectively, with filter size 3, stride length 1). The network applies ReLU activation to the output feature maps, with a max-pooling layer after the first two convolutional layers. The final convolutional layer output is flattened, followed by an $L_{2}$ normalized dense layer.

We use multi-similarity loss~\cite{wang2019multi} which uses two iterative steps: pair-mining and weighting. This approach considers both self-similarity and relative similarity to collect more informative pairs, and takes a weighted combination of selected positive and negative pairs. Like other pairwise-based losses, this loss function maximizes the distance between dissimilar examples and minimizes it between similar examples.

Equation~\ref{eq1} provides the formulation of the multi-similarity loss function: 

\vspace*{-2mm}
\begin{multline} \label{eq1}
   \frac{1}{m} \sum_{i=1}^{m} \{ \frac{1}{\alpha} \log[1 + \sum_{k \in P_{i}} 1 + e^{- \alpha (S_{ik}- \lambda)}] \\
   + \frac{1}{\beta} \log[1 + \sum_{k \in N_{i}} 1 + e^{\beta (S_{ik}- \lambda)}] \},
\end{multline}
\vspace*{-2mm}

where $N_{i}$ represents negative pairs (samples from different classes) in the batch while $P_{i}$ denotes positive pairs (samples from the same class). $S_{ik}$ represents element $(i,k)$ of the similarity matrix, indicating the similarity of two samples $\{x_{i}, x_{k}\}$, $S_{ik}\text{ }:=\text{ }f(x_{i};\theta)\cdot f(x_{k};\theta)$ where $f$ is the neural network with parameters $\theta$. The cosine embedding size is 64.

We use Adam optimization~\cite{DBLP:journals/corr/KingmaB14} with a learning rate of $5 \times 10^{-6}$, with batch size 70, and train for 20 epochs. Training was performed on a Mac M1 Max with Metal acceleration. In every mini-batch, 10 inputs ($m=10$) from each of the 7 training classes are randomly sampled. In Equation~\ref{eq1}, $\alpha = 2$ (weight for positive pairs), $\beta = 40$ (weight for negative pairs), $\lambda = 0.5$ (used to weight the distance). Margin $\epsilon = 0.1 $ is used to remove easy positive and negative pairs such that negative pairs are sampled if they are greater than ($\underset{y_{i} = y_{k}}\min(S_{ik}) - \epsilon$) where {$\underset{y_{i} = y_{k}}\min S_{ik}$} represents the positive pair with the lowest similarity, and positive pairs are sampled if they are less than ($\underset{y_{i} \neq y_{k}}\max(S_{ik}) + \epsilon$) where $\underset{y_{i} \neq y_{k}}\max S_{ik}$ represents the negative pair with the highest similarity. $y$ denotes the one-hot label vectors. 

Since during training only {\it 7} types of flat-sided and round objects are used, the model learns to output an embedding that represents pure round and flat objects samples in the cosine space. The extracted embeddings are indexed. Given that the index of the embedding space represents only purely round or flat objects, we consider 100 test samples each from all {\it 11} classes (seen and unseen) and find the closest matches to the test samples using a nearest neighbour search ($K=10$). 

\vspace*{-2mm}
\paragraph{Similarity Learning Results}

\begin{figure}
    \centering
    \includegraphics[width=.5\textwidth, trim=0px 20px 50px 0px, clip]{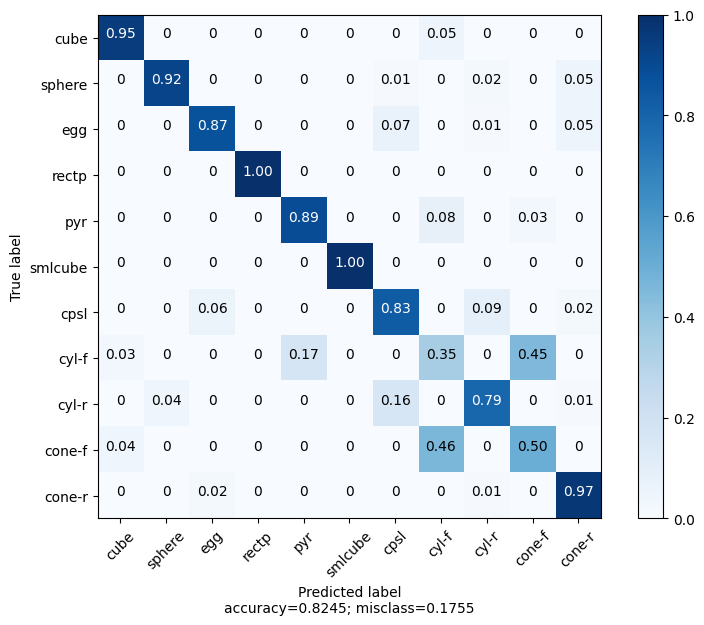}
    \vspace*{-2mm}
    \caption{Confusion matrix on the test split of 11 objects. Only 7 pure flat and round objects are used during training. {\tt cyl-f} = cylinder, flat side down; {\tt cyl-r} = cylinder, round edge down; likewise for {\tt cone-f}/{\tt r}. The values shown in the matrix are normalized between 0 and 1.}
    \label{fig:cm}
    \vspace*{-2mm}
\end{figure}

Fig.~\ref{fig:cm} shows the confusion matrix for nearest neighbor search on the test split of objects, using 100 test samples per class. Interestingly, even though the model was not trained on any \textit{cone} and \textit{cylinder} instances, it is still able to not only match them to the correct object type, but also to the correct orientation. Where confusions arise, it is between different flat-sided objects and different round objects, but never across these categories.  In other words, this model can capture and distinguish the main distinguishing concepts---roundness and flatness---in the different object classes, and draw comparisons between them across classes. It also applies what is already learned to novel objects to find similar examples with respect to these concepts. Overall classification accuracy is 82\%.

\vspace*{-2mm}
\subsection{Grounding Conceptual Vocabulary}
\label{ssec:grounding}
\vspace*{-1mm}

\begin{figure}
    \centering
    \includegraphics[width=.5\textwidth]{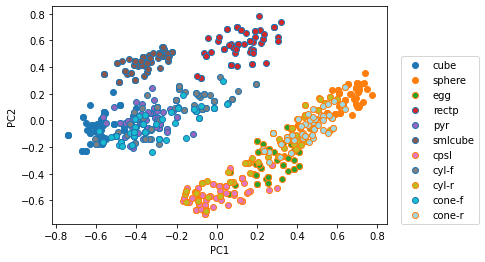}
    \vspace*{-2mm}
    \caption{PCA of test object embeddings. Points outlined in blue represent ``flat'' object embeddings. Points outlined in orange represent ``round'' object embeddings.}
    \label{fig:object-pca}
    \vspace*{-2mm}
\end{figure}

First, we extracted the embeddings of 800 object test samples from the learned object space.  These were 64D embeddings that defined the object representation space, and objects clustered into two broad regions defining the ``flat-sided'' and ``round-sided'' objects (see Fig.~\ref{fig:object-pca}).

Individual embedding vectors of different instances of the same object type form a region defining the object representation where some subset of these vectors form the region's spanning set; ~\citet{ethayarajh-2019-contextual} observed similar phenomena in the representations of contextualized token vectors from LLMs, suggesting there exists a structure-preserving mapping between equivalent regions in different embedding spaces.

To assess this, we needed to generate appropriately contextualized vector representations of terms to ground to the object representation space.  For this we turned to OpenAI's ChatGPT model to rapidly generate a sentence corpus. ChatGPT was given prompts to generate short, unique sentences that would explicitly mention the objects by name and describe their behavior in a stacking task (e.g., ``{\it Write 40 short sentences about how cubes can be stacked}''). In the process, ChatGPT also generated mentions of properties of the objects ({\it flat/round}), associated behaviors ({\it stack/roll}), properties of the resulting structure ({\it stable/unstable}), and resulting state of the structure ({\it stand/fall}). We generated 40 sentences describing each object type, plus 20 sentences each for {\it block} and {\it ball}, synonyms for {\it cube} and {\it sphere}. In total, a 440-sentence corpus was generated.

We then took the most frequently-occurring domain-relevant terms (these were the object names and aforementioned related conceptual terms) and extracted the word-level embeddings for each occurrence. We extracted word embeddings using the BERT, RoBERTa, ALBERT (all 768D), and XLM (2,048D) base models. Embeddings were creating by summing over the encoder hidden states of the last four encoder layers.  Where tokenization split the target word into multiple tokens, the individual contextualized token embeddings were averaged to create a single embedding.

To actually ground the word embeddings into the object space, we used a simple affine transformation. We took 5 contextualized embeddings of each target word, paired each with an embedding for the object whose name occurs in the sentence the target word came from, and use them to compute an affine transformation from LLM space to object embedding space, using a ridge regressor that minimizes the mean squared-error distance between the paired embeddings. The resulting transformation matrix serves as a ``bridge'' between the two representation spaces.  This affine transformation technique has previously been used to compare image embeddings from different vision models~\cite{mcneely2022canonical} and to map information from monolingual LLMs into multilingual LLMs~\cite{nath2022phonetic}. Here we apply this technique in a cross-modal setting.

We perform iterative experiments, starting by using only a subset of the different words and objects to compute the mapping, and incrementally add conceptual vocabulary to improve the quality of the calculated transformation. We evaluate the transformation by transforming word vectors for concepts not used in computing the transformation matrix and seeing if those embeddings cluster with the correct set of objects that bear those properties, have those affordances, etc. The order in which object concepts are introduced follows the order we used previously in \citet{ghaffari2022detecting}, with the exception of moving {\it cylinder} and {\it cone} to the end, due to their exclusion from initial training of the similarity learning model, and pairing one flat-sided with one round object (e.g., {\it pyramid} + {\it capsule}) at each step.

As a final step, a ``hint'' is provided by adding 5 embedding pairs that explicitly include the concept to be grounded to the computation of the transformation. We evaluate this by transforming new instances of that concept into the object embedding space and seeing where they cluster.  We quantify the clusters of different concepts when transformed into the object space using separation of cluster centers and a K-nearest neighbor (KNN) classifier with $K=5$.

\vspace*{-2mm}
\section{Results of Conceptual Grounding}
\label{sec:results}
\vspace*{-2mm}

\begin{figure*}
    \centering
    \includegraphics[height=1.15in, trim=10px 0px 95px 0px, clip]{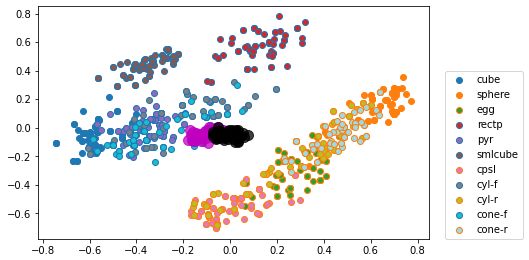}
    \includegraphics[height=1.15in, trim=10px 0px 95px 0px, clip]{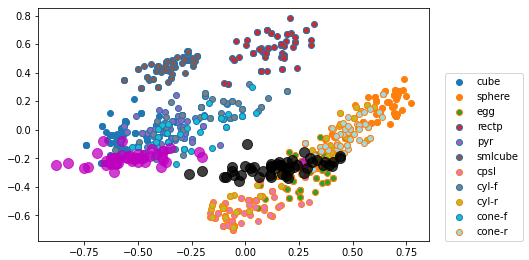}
    \includegraphics[height=1.15in, trim=10px 0px 5px 0px, clip]{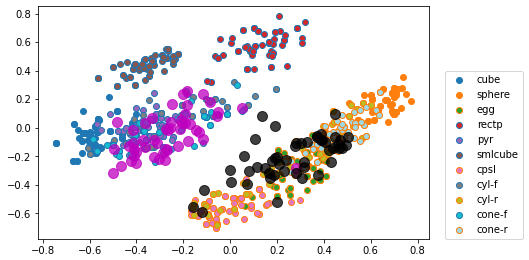}
    \vspace*{-2mm}
    \caption{PCA of ``flat'' (pink) and ``round'' (black) test embeddings from XLM mapped into object embedding space. L: with mapping computed using only information about {\it cubes}, {\it spheres}, and {\it eggs}. C: using information about all objects. R: using all objects and a 5-sample ``hint'' about ``flat'' vs. ``round.''}
    \label{fig:flat-round}
    \vspace*{-2mm}
\end{figure*}

For illustrative purposes, let us first examine the concepts of ``flat'' and ``round'' using word embeddings drawn from XLM, the best-performing model in this domain when ``hinting'' is used.  Further results from other models are given in the Appendix.

In Fig.~\ref{fig:flat-round}, we see word embeddings for ``flat'' and ``round'' transformed into the object embedding space.  At first (the left figure), when only information about cubes, spheres, and eggs are used to compute the mapping, there is only a slight separation between the two transformed embedding clusters and neither term clusters cleanly with either flat or round objects. When information about all objects is used to compute the transformation (center), the two word embedding clusters distinctly separate, with most ``flat'' embeddings clearly overlapping with the flat-sided objects, {\it mutatis mutandis} ``round'' embeddings and the round objects. Finally (right), the ``hint'' is provided, by explicitly pairing a small set of 5 ``flat'' or ``round'' word embeddings to object embeddings whose type appears in a generated sentence collocated with the target word (e.g., ``{\it The \underline{cubes} were \underline{flat} on all sides, making it easy to stack them neatly.}''). With this hint we see that the ``flat'' and ``round'' word embeddings more completely overlap with the objects that have the respective attributes.

\begin{figure}[h!]
    \centering
    \includegraphics[width=.45\textwidth]{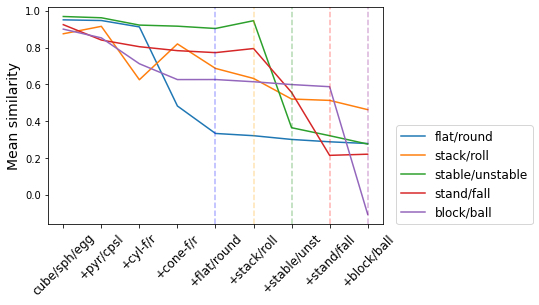}
    \includegraphics[width=.45\textwidth]{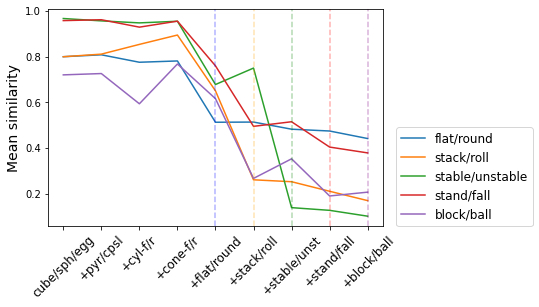}
    \includegraphics[width=.45\textwidth]{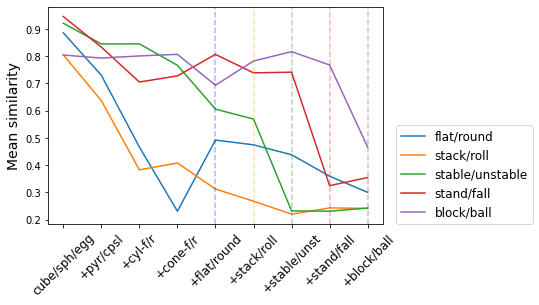}
    \includegraphics[width=.45\textwidth]{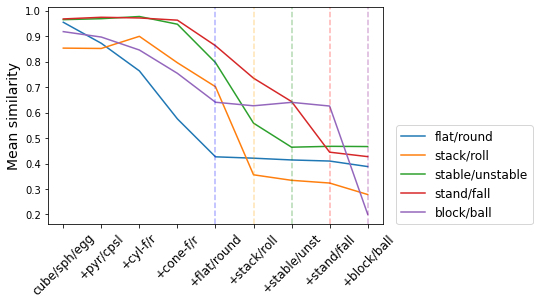}
    \vspace*{-2mm}
    \caption{Separation of cluster centers for transformed (in order) BERT, RoBERTa, ALBERT, and XLM embeddings for verb and property concepts, as more information about other concepts is progressively added to compute the transformation. Dashed lines show where a ``hint'' is given about the concept to be grounded (denoted by the similarly-colored solid line).}
    \label{fig:meanseps-objfirst}
    \vspace*{-4mm}
\end{figure}

\begin{figure}[h!]
    \centering
    \includegraphics[width=.45\textwidth]{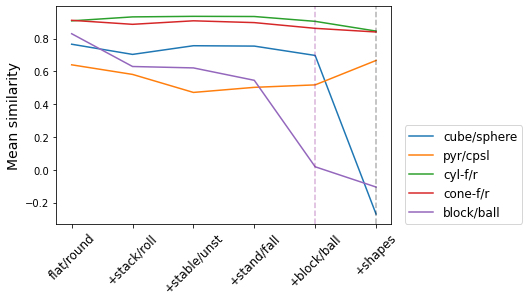}
    \includegraphics[width=.45\textwidth]{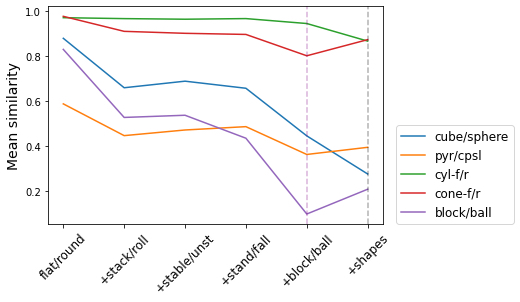}
    \includegraphics[width=.45\textwidth]{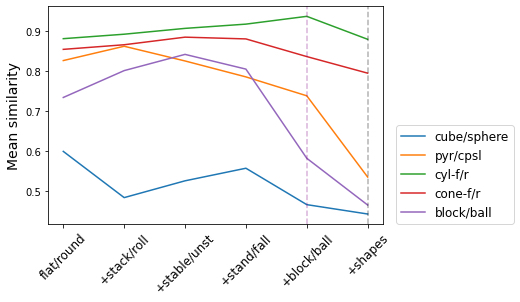}
    \includegraphics[width=.45\textwidth]{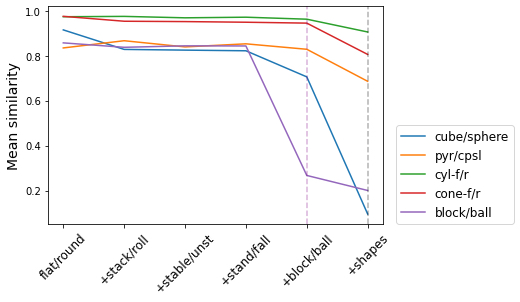}
    \vspace*{-2mm}
    \caption{Separation of cluster centers for transformed embeddings for object concepts, as more information about other concepts is progressively added to compute the transformation. Format is identical to Fig.~\ref{fig:meanseps-objfirst}. {\tt +shapes} denotes adding information about all objects.}
    \label{fig:meanseps-absfirst}
    \vspace*{-4mm}
\end{figure}

When little information about related object concepts is provided when computing the mapping from LLM space to object embedding space, the transformed clusters of contrasting terms share a high level of similarity in object space, but as more information about related object concepts is introduced into the transformation, the separation of the transformed novel concept clusters start to cleanly separate and become distinguished from each other.  Fig.~\ref{fig:meanseps-objfirst} shows the mean similarity of the transformed clusters of attributive, verbal, and object synonym terms as different object terms are mapped into the object space, using embeddings drawn from the four different LLMs.  Fig.~\ref{fig:meanseps-absfirst} shows the same change in the similarity between cluster centers, but this time evaluated over the transformed {\it object} terms when the transformation is computed using attributive and verbal terms. In both plots, dashed lines show where an explicit ``hint'' is given about specific concepts. We see that just by using a few samples of each concept and projecting them into object space using an affine transformation, grounding object terms is helpful in distinguishing the meaning of terms denoting related properties, attributes, and verbs, but grounding the more abstract concept vocabulary first does not usually cause the transformed clusters of object terms to separate before explicit hinting is provided, reflecting the psycholinguistic hypothesis of \citet{gentner1983structure}.

Table~\ref{tab:final_results1} shows the results of the KNN classifier over the transformed attributive and verbal word embeddings, both when the transformation was computed using only object information (top section) and with ``hints'' about the attributive concepts (bottom).  Table~\ref{tab:final_results2} shows KNN classifier results over transformed object embeddings without hints about the objects, and with.  We report macroaveraged F1 scores, so that successful performance on high support classes does not obscure poor performance on low support classes.  Numbers in parentheses show how much ``hinting'' helped improve performance of the particular model on the concept in question.  {\it Block} and {\it ball} are included in both the ``object'' test set and the ``predicate'' test set (even though they are not predicative terms in this sense), because these terms were not used in computing the affine mapping in either case. They are included as synonyms for {\it cube} and {\it sphere}. Further discussion is provided in Sec.~\ref{sec:disc}.

\begin{table*}[h!]
    \centering
    \begin{tabular}{rlllll}
        \toprule
        & \small{{\bf flat/round}} & \small{{\bf stack/roll}} & \small{{\bf stable/unstable}} & \small{{\bf stand/fall}} & \small{{\bf block/ball}}  \\
        \small{{\bf Models}} & \small{$N=103$} & \small{$N=56$} & \small{$N=22$} & \small{$N=10$} & \small{$N=30$}  \\
        \cmidrule(lr){1-6}
        \small{BERT} & \small{0.89}& \small{0.16} &\small{\bf 0.58} &\small{0.60} &\small{0.33}\\
        \small{RoBERTa} & \small{0.34}& \small{0.16} &\small{0.29} &\small{0.37} &\small{0.67}\\
        \small{ALBERT} & \small{\bf 0.92}& \small{\bf 0.65} &\small{\bf 0.58} &\small{\bf 0.89} &\small{0.60}\\
        \small{XLM} & \small{0.73}& \small{0.53} &\small{0.37} &\small{0.29} &\small{\bf 0.79}\\
        \cmidrule(lr){1-6}
        \small{BERT+hint} & \small{0.96 (+0.07)}& \small{0.78 (+0.62)} &\small{0.91 (+0.63)} &\small{\textbf{1.00} (+0.40)} &\small{0.93 (+0.60)}\\
        \small{RoBERTa+hint} & \small{0.90 (+0.56)}& \small{0.89 (+0.73)} &\small{\textbf{1.00} (+0.71)} &\small{\textbf{1.00} (+0.63)} &\small{0.90 (+0.23)}\\
        \small{ALBERT+hint} & \small{0.89 (-0.03)}& \small{0.85 (+0.20)} &\small{0.86 (+0.28)} &\small{\textbf{1.00} (+0.11)} &\small{0.66 (+0.06)}\\
        \small{XLM+hint} & \small{\textbf{0.98} (+0.25)}& \small{\textbf{1.00} (+0.47)} &\small{0.73 (+0.36)} &\small{\textbf{1.00} (+0.71)} &\small{\textbf{0.97} (+0.18)}\\
        \cmidrule(lr){1-6}
        \bottomrule
    \end{tabular}
    \vspace*{-2mm}
    \caption{Macroaveraged KNN F1 over transformed attribute/verb/synonym word embedding test sets (mapping computed using object embeddings). Numbers in parentheses show performance increase with ``hinting.''}
    \label{tab:final_results1}
\vspace*{-2mm}
\end{table*}

\begin{table*}[h!]
    \centering
    \begin{tabular}{rlllll}
        \toprule
        \small{{\bf Models}} & \small{{\bf cube/sphere}} & \small{{\bf pyr/cpsl}} & \small{{\bf cyl-f/r}} & \small{{\bf cone-f/r}} & \small{{\bf block/ball}}  \\
        \cmidrule(lr){1-6}
        \small{BERT} & \small{0.77}& \small{0.46} &\small{0.34} &\small{0.40} &\small{\bf 0.83}\\
        \small{RoBERTa} & \small{0.81}& \small{0.44} &\small{0.40} &\small{0.49} &\small{0.55}\\
        \small{ALBERT} & \small{\bf 0.88}& \small{\bf 0.88} &\small{\bf 0.81} &\small{\bf 0.78} &\small{0.46}\\
        \small{XLM} & \small{0.40}& \small{0.46} &\small{0.49} &\small{0.36} &\small{0.55}\\
        \cmidrule(lr){1-6}
        \small{BERT+hint} & \small{0.97 (+0.20)}& \small{\textbf{1.00} (+0.54)} &\small{0.78 (+0.44)} &\small{0.84 (+0.44)} &\small{0.93 (+0.10)}\\
        \small{RoBERTa+hint} & \small{0.81 ($\pm$0.00)}& \small{0.94 (+0.50)} &\small{0.78 (+0.38)} &\small{0.87 (+0.38)} &\small{0.90 (+0.35)}\\
        \small{ALBERT+hint} & \small{0.88 ($\pm$0.00)}& \small{0.94 (+0.06)} & \small{\textbf{0.87} (+0.06)} &\small{0.88 (+0.10)} &\small{0.66 (+0.20)}\\
        \small{XLM+hint} & \small{\textbf{1.00} (+0.60)}& \small{0.97 (+0.51)} &\small{0.81 (+0.32)} &\small{\textbf{0.91} (+0.55)} &\small{\textbf{0.97} (+0.42)}\\
        \cmidrule(lr){1-6}
        \bottomrule
    \end{tabular}
    \vspace*{-2mm}
    \caption{Macroaveraged KNN F1 over transformed object word embedding test sets (mapping computed using attribute/verb embeddings). Numbers in parentheses show performance increase with ``hinting.'' $N=30$ for all.}
    \label{tab:final_results2}
\vspace*{-2mm}
\end{table*}

\vspace*{-2mm}
\section{Discussion}
\label{sec:disc}
\vspace*{-2mm}

\paragraph{Separation of conceptual clusters}

In Fig.~\ref{fig:meanseps-objfirst}, we can see that for object concept vectors from certain models, as information about certain other concepts is included in the transformation from LLM space to object space, the centers of the conceptual clusters in question start to organically separate. This is particularly true for ALBERT object word vectors and to some extent XLM and BERT vectors.  In other words, if the model already ``knows'' about the dual aspects of cones and cylinders, it becomes easier to distinguish an abstract concept of {\it flatness} from {\it roundness}.  Clusters of transformed RoBERTa object word vectors tend not to separate very clearly until explicit hints about them are provided.

{\it Flat/round} is the easiest of the attributive or verbal concepts to distinguish, through affine transformations that include information about flat and round objects.  {\it Stable/unstable} is a particularly hard distinction for most model representations, in part because of the low support for these terms in the training corpus but also because in the scenario captured in the simulation data and described in training sentences, the terms refer to properties not of the objects themselves, but of the objects in the context of the stacking task (i.e., spheres are not inherently ``unstable'' but are if someone attempts to stack them). This suggests data gathered from either stacking more objects, or from tasks involving more complex balancing acts would be useful to learn a robust interpretation of such terms.

Inverse trends are observable in Fig.~\ref{fig:meanseps-absfirst}, where we see that when the transformation includes only information about attributes and verbs, transformed BERT and XLM object word vectors for contrasting objects do not meaningfully separate until explicit hints are provided (and even then sometimes they don't separate much). Some of the RoBERTa object word clusters do appear to appreciably separate as more attribute and verb information is added to the transformation, and ALBERT object word clusters, actually at first grow {\it closer} as related conceptual information is added to the transformation, until suddenly separating at the provision of an explicit hint. This suggests that ALBERT, perhaps due to its smaller training size and architecture, learns vocabulary representations that are more ``entangled'' or that representations of flat-sided or round object words carry with them a bias toward object-related interpretations of ``flat,'' ``round,'' and associated terms.  Meanwhile XLM and other representations of abstract vocabulary are perhaps less correlated with concrete nouns, making them less easy to ground but also in principle more compositional with less bias toward certain interpretations.

\vspace*{-2mm}
\paragraph{Classification of conceptual terms}

With hinting, XLM vectors perform best in the term classification task.  XLM is the largest of the four models and has the largest embedding size (2,048 where all other models use an embedding size of 768).  Hinting typically provides the biggest boost in performance to XLM vectors, both for grounding concrete object and abstract terms.  This suggests that the object concepts and the attributive/verbal concepts form distinct and possibly distant regions in the original XLM embedding space, and that an affine transformation into the object space does not always put pairs of contrastive attributes or verbs closer to distinct objects that display those respective properties.  Providing hints helps all models achieve high performance by matching objects and related concepts, but the performance boost is particularly high for XLM vectors, which often perform very badly in KNN classification of some concepts (e.g., {\it stable/unstable}, {\it stand/fall}) until hints are provided. Hinting is still less helpful for transforming XLM object vectors when only previous information about attributes or verbs is provided.

Interestingly, hinting is least helpful when grounding word vectors from ALBERT, the smallest of the four models. On eight out of ten concept pairs explored, ALBERT vectors perform the best by far in the KNN classifier before any hints are provided, but providing subsequent hints makes only a small difference to classification F1, and sometimes none at all, while boosting the performance of other representation ahead of ALBERT vectors. This suggests that the object and related concept representations already share some level of correlation and possibly overlap in ALBERT embedding space. In turn, these results suggest that larger models like XLM may be better able to represent multiple word senses and figurative, non-physically-grounded usages of terms like these. However, grounding these concepts to a physical environment without some explicit ``nudges'' may be more challenging for larger pretrained models than smaller ones, in which the abstract concepts may already be biased toward correlations with the associated concrete object concepts. Further discussion is provided in the Appendix.

\vspace*{-2mm}
\section{Conclusion and Future Work}
\label{sec:conc}
\vspace*{-2mm}

In this paper, we have presented evidence that similarity learning over rich object behavior and trajectory data from an embodied simulation environment can create a representation space that not only successfully classifies concrete objects but can make analogical comparisons between them based on abstract properties that inhere across multiple object types. We used the resulting representation to conduct investigations into the properties of token embeddings from different LLMs by mapping them into the object space using a linear ridge regression technique. We found that computing a mapping using representations of objects/object terms correlated with increased ability to distinguish and assign related conceptual vocabulary to the right categories, but that representations from different LLMs behaved quite differently. We also observed that computing mappings using information about abstract properties was less useful for distinguishing and classifying object terms. This reflects earlier arguments from psycholinguistics and analogical reasoning, e.g., \citet{gentner2006verbs}'s hypothesis that names for concrete objects should be learnable by humans very early but that associated verbs and attributes are harder. 

Our approach uses numerical data that situates and embodies an agent's positioning in the environment relative to the objects it interacts with. This method allows us to build a model over rich information without visual artifacts like occlusion or perspective distortion,  Prior research, e.g., \citet{krishnaswamy2022affordance,pustejovsky2022multimodal} has demonstrated that embodiment is also influenced by other factors like events and habitats, and that purely linguistic representations of objects, attributes, and activities may not capture these types of information. In fact, the corpus generated using ChatGPT, an unembodied language model trained solely over text, is likely not entirely representative of these aspects beyond cooccurrences between object terms, habitats, and affordances in the training data.  What our embodied approach brings is a way to correlate representations extracted from unembodied models to representations learned from embodied data, and provides evidence that the ability to ground real-world entities, properties, or actions to lexical items could enable LLMs to simulate the human ability to link utterances to specific communicative intents.  However, further investigation is necessary.

Since the primary objective of this research is to provide a method that achieves human-like ``understanding'' of communicative intents, we should note that we do not argue that human learners use the same mathematical transformations we use herein, but just that we can use them to make AI systems behave similarly.

Directions for future work include
\begin{enumerate*}[label=\arabic*)]
    \item investigating the effects of intra-class order when grounding tokens, e.g., introducing object concepts to the affine mapping in a different order;
    \item using similarity learning over images, or images combined with the embodied data, to create the representation space;
    \item using data gathered in other embodied tasks to investigate other concepts like concavity or directedness, that are not captured in this stacking task;
    \item evaluating token representations directly from a decoder like a GPT-style model; and
    \item directly operationalizing analogical comparisons in a real-time embodied simulation, e.g., by making an agent solve problems using analogical reasoning in a live environment.
\end{enumerate*}

\section*{Acknowledgments}

We would like to thank our reviewers for their helpful suggestions on improving the final copy of this paper.  This research was supported in part by a grant from the U.S. Army Research Office (ARO) to Colorado State University, under award \#W911NF-23-1-0031. The positions expressed herein do not reflect the official position of the U.S. Department of Defense or the United States government. Any errors or omissions are the responsibility of the authors.


\bibliographystyle{acl_natbib}
\bibliography{anthology,acl2021}

\clearpage
\appendix

\section{Appendix: Additional Results}

The following figures are provided for comparison with Fig.~\ref{fig:flat-round} and Table~\ref{tab:final_results1}.

\begin{figure}[h!]
    \centering
    \includegraphics[width=.5\textwidth]{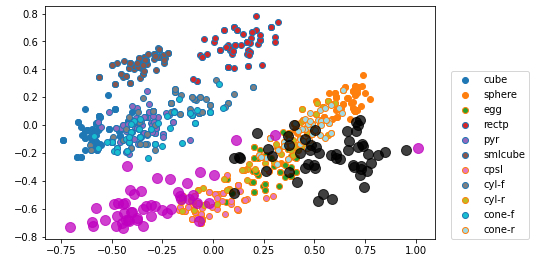}
    \caption{PCA of ``flat'' (pink) and ``round'' (black) test embeddings from BERT mapped into object representation space. Mapping is computed using information about all objects but without flat/round hinting.}
    \label{fig:bert-flatround-nohint}
    \vspace*{-4mm}
\end{figure}

\begin{figure}[h!]
    \centering
    \includegraphics[width=.5\textwidth]{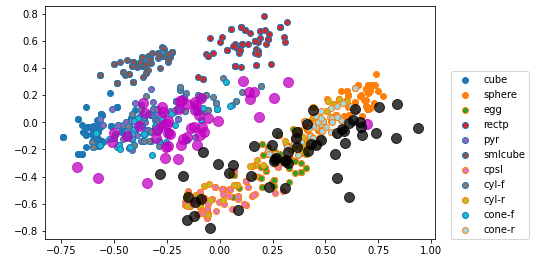}
    \caption{PCA of ``flat'' (pink) and ``round'' (black) test embeddings from BERT mapped into object representation space. Mapping is computed using information about all objects and flat/round hinting.}
    \label{fig:bert-flatround-hint}
    \vspace*{-2mm}
\end{figure}

Fig.~\ref{fig:bert-flatround-nohint} shows the projection of ``flat'' and ``round'' token embeddings from \textbf{BERT} into the learned object representation space when the mapping is computed using paired embeddings of objects and object terms, but no explicit hint is provided about the meaning of ``flat'' and ``round.''  The two clusters clearly separate from each other but do not map clearly onto flat and round object representations at this stage.  Fig.~\ref{fig:bert-flatround-hint} shows the same projection after a 5-sample hint about ``flat'' and ``round'' is added to the mapping.

\begin{figure}[h!]
    \centering
    \includegraphics[width=.5\textwidth]{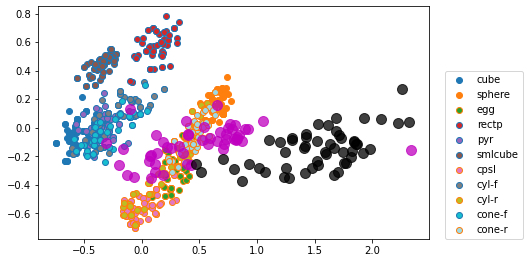}
    \caption{PCA of ``flat'' (pink) and ``round'' (black) test embeddings from RoBERTa mapped into object representation space. Mapping is computed using information about all objects but without flat/round hinting.}
    \label{fig:roberta-flatround-nohint}
    \vspace*{-4mm}
\end{figure}

\begin{figure}[h!]
    \centering
    \includegraphics[width=.5\textwidth]{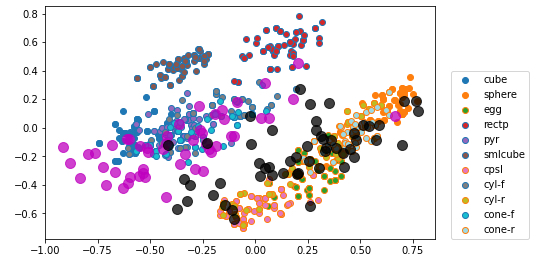}
    \caption{PCA of ``flat'' (pink) and ``round'' (black) test embeddings from RoBERTa mapped into object representation space. Mapping is computed using information about all objects with flat/round hinting.}
    \label{fig:roberta-flatround-hint}
    \vspace*{-2mm}
\end{figure}

Fig.~\ref{fig:roberta-flatround-nohint} shows the projection of ``flat'' and ``round'' token embeddings from \textbf{RoBERTa} into the learned object representation space when the mapping is computed using paired embeddings of objects and object terms, but no explicit hint is provided about the meaning of ``flat'' and ``round.''  Again, the two clusters clearly separate from each other at this stage, but the ``flat'' embeddings are closer to the round objects embeddings in the 64D space while the ``round'' embeddings are distinct from each object cluster.  Fig.~\ref{fig:roberta-flatround-hint} shows the same projection after a 5-sample hint about ``flat'' and ``round'' is added to the mapping.

\begin{figure}[h!]
    \centering
    \includegraphics[width=.5\textwidth]{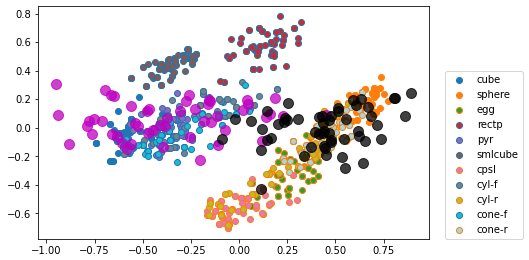}
    \caption{PCA of ``flat'' (pink) and ``round'' (black) test embeddings from ALBERT mapped into object representation space. Mapping is computed using information about all objects but without flat/round hinting.}
    \label{fig:albert-flatround-nohint}
    \vspace*{-2mm}
\end{figure}

\begin{figure}[h!]
    \centering
    \includegraphics[width=.5\textwidth]{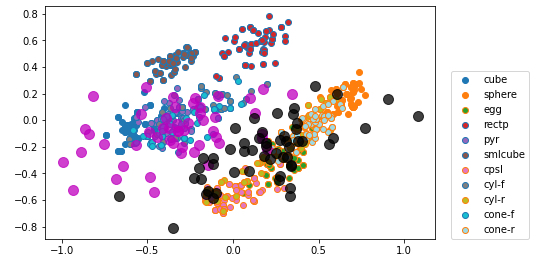}
    \caption{PCA of ``flat'' (pink) and ``round'' (black) test embeddings from ALBERT mapped into object representation space. Mapping is computed using information about all objects with flat/round hinting.}
    \label{fig:albert-flatround-hint}
    \vspace*{-2mm}
\end{figure}

Figs.~\ref{fig:albert-flatround-nohint} and \ref{fig:albert-flatround-hint} show the equivalent using the \textbf{ALBERT} ``flat''/``round'' token embeddings.  Here, without hinting, the transformed ``flat'' embeddings mostly cluster with flat-sided objects and the transformed ``round'' embeddings mostly cluster with round objects, suggesting that in ALBERT, the representations of ``flat'', ``round'', and other associated object-related concepts are relatively entangled with the object terms themselves.  Hinting solidifies this correlation somewhat but the effect is relatively small, as discussed in Sec.~\ref{sec:disc}.

\begin{figure}[htb]
    \centering
    \includegraphics[width=.5\textwidth]{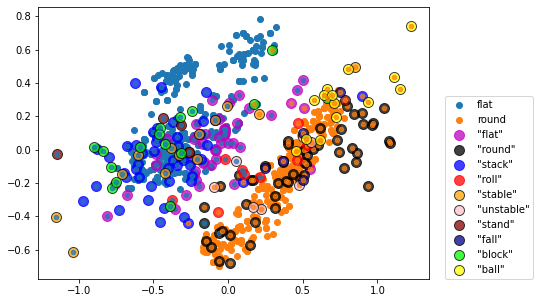}
    \includegraphics[width=.5\textwidth]{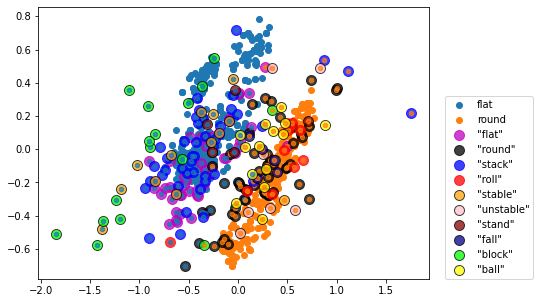}
    \includegraphics[width=.5\textwidth]{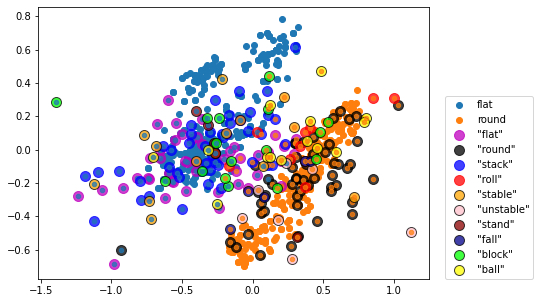}
    \includegraphics[width=.5\textwidth]{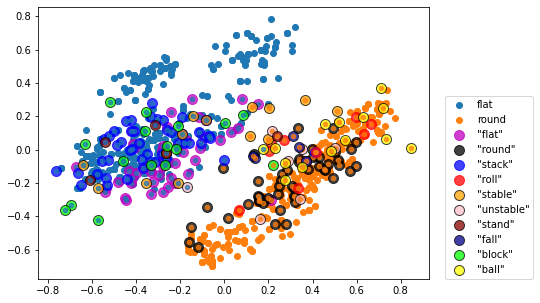}
    \caption{PCA of (top to bottom) BERT, RoBERTa, ALBERT, and XLM test word embeddings for all concepts mapped into object representation space, including hinting in the mapping. Innermost colored point indicates whether that transformed embedding clusters with flat-sided objects or round objects.}
    \label{fig:fullplots}
    \vspace*{-4mm}
\end{figure}

Fig.~\ref{fig:fullplots} shows contextualized token embeddings for {\it all} vocabulary items from (top to bottom) BERT, RoBERTa, ALBERT, and XLM mapped into the object representation space when the mapping is computed using information about all concepts, including hinting.  For all points, the outer color denotes the token it represents and the inner color (blue or orange) indicates whether the transformed embedding clusters with flat-sided or round object representations. Therefore, a black point with an orange center indicates a ``round'' token embedding that clusters with round objects (correctly), but a black point with a {\it blue} center indicates one that incorrectly clusters with flat-sided objects.  We see that when using the full set of concepts in computing the mapping between spaces, the larger models show the strongest correlations between correctly-mapped token embeddings and the expected set of object representations.  Mapped XLM vectors show the strongest separation between the flat-related concepts and round-related concepts, while mapped ALBERT vectors display a fairly significant overlap between those correlated with flat objects and those correlated with round object (this is evident in the center of the plot ``between'' the two main flat and round clusters).  Mapped RoBERTa and to a lesser extent BERT embeddings show a similar overall separation to mapped XLM embeddings, but with a wider dispersion in the distribution of mapped embeddings, where some (particularly in the case of RoBERTa embeddings) have a very high Euclidean distance from the two core object representation clusters to which they are compared.

\end{document}